\documentclass[a4paper, 10pt, conference]{ieeeconf}
\IEEEoverridecommandlockouts

\usepackage{xspace}
\usepackage{amsmath}
\usepackage{algorithm}
\usepackage[noend]{algorithmic}
\usepackage[caption=false,font=footnotesize]{subfig}	
\usepackage{framed} 
\usepackage[]{epsfig,color}
\usepackage[nice]{nicefrac}
\usepackage{multirow}
\usepackage{url}
\usepackage{wrapfig}
\usepackage[switch, columnwise]{lineno}

\newcommand{\Cpp}{C\raise.08ex\hbox{\tt ++}\xspace}

\newcommand{\Cfree}{\ensuremath{\calX_{\text{free}}}\xspace}
\newcommand{\Cforb}{\ensuremath{\calX_{\text{forb}}}\xspace}
\newcommand{\Cs}{C-space\xspace}
\newcommand{\Css}{C-spaces\xspace}

\newcommand{\kdtree}{$kd$-tree\xspace}
\newcommand{\kdtrees}{$kd$-trees\xspace}
\newcommand{\nn}{nearest-neighbor\xspace}
\newcommand{\nns}{nearest-neighbors\xspace}
\newcommand{\rtg}{RTG\xspace}
\newcommand{\MPd}{MP\xspace}	

\def\naive{{na\"{\i}ve}\xspace}

\newcommand{\set}[1]{\ensuremath{\{ #1\}}}

\newcommand{\calG}{\ensuremath{\mathcal{G}}\xspace}

\newcommand{\calX}{\ensuremath{\mathcal{X}}\xspace}

\newcommand{\calT}{\ensuremath{\mathcal{T}}\xspace}

\newcommand{\ignore}[1]{}

\newboolean{ICRA}
\newboolean{ARXIV}

\setboolean{ICRA}{false}
\ifthenelse{\boolean{ICRA}}
	{\setboolean{ARXIV}{false} }
	{\setboolean{ARXIV}{true}  }
\newcommand{\textVersion}[2]
{\ifthenelse{\boolean{ICRA} }{#1}{}\ifthenelse{\boolean{ARXIV}}{#2}{}}

\textVersion
{
\setlength\floatsep{0.5\baselineskip plus 3pt minus 2pt}
\setlength\textfloatsep{0.5\baselineskip plus 3pt minus 2pt}
}
{
}
\begin{document}
\pagestyle{plain}
\pagenumbering{arabic}

%

\title{	\LARGE \bf
        Efficient high-quality motion planning by\\
        fast all-pairs $r$-nearest-neighbors
			}
\author{Michal Kleinbort, Oren Salzman and Dan Halperin$^*$
\thanks{
$^*$
Blavatnik School of Computer Science,
Tel-Aviv University, Israel}
\thanks{
This work has been supported in part by the
Israel Science Foundation (grant no. 1102/11),
by the German-Israeli Foundation (grant no. 1150-82.6/2011), and
by the Hermann Minkowski--Minerva Center for Geometry at Tel Aviv
University.}}%

\maketitle


\begin{abstract}
Sampling-based motion-planning algorithms typically rely on nearest-neighbor (NN) queries when constructing a roadmap.
Recent results 
suggest that in various settings NN queries may be the computational bottleneck of such algorithms.
Moreover, in several \emph{asymptotically-optimal} algorithms
these NN queries are of a specific form:
Given a set of points and a radius $r$
report all pairs of points whose distance is at most $r$.
This calls for an application-specific NN data structure tailored to efficiently answering this type of queries.
Randomly transformed grids (\rtg) were recently proposed by Aiger et al.~\cite{AKS13} as a tool to answer such queries
and have been shown to outperform common implementations of NN data structures in this context.
In this work we employ \rtg
for sampling-based motion-planning algorithms and
describe an efficient implementation~of the approach.
We show that for motion-planning,
\rtg allow for
faster convergence to high-quality solutions when compared with existing NN  data structures.
Additionally, \rtg enable significantly shorter construction times
 for batched-PRM variants;
specifically,
we demonstrate a speedup by a factor of two to three for some scenarios.

\end{abstract} 
\section{Introduction and Related work}
Given a robot moving in an environment cluttered with obstacles, motion-planning (MP)
algorithms are used to efficiently plan a path for the robot, while avoiding collision with the obstacles~\cite{CBHKKLT05}.
A common approach is to use \emph{sampling-based} algorithms,
which abstract the robot as a point in a high-dimensional space called the \emph{configuration space} (\Cs) and plan a path in this space.
A point, or a
configuration, in the \Cs represents a placement of the robot that is either collision-free or not,
subdividing the \Cs~$\calX$ into the sets \Cfree and \Cforb, respectively.
The structure of the \Cs is then studied by constructing a graph, called a \emph{roadmap},
that approximates the connectivity of~\Cfree.
The nodes of the graph are collision-free configurations sampled at random.
Two (nearby) nodes are connected by an edge if the straight line connecting their configurations is
collision-free as well.

Sampling-based \MPd algorithms are implemented using two  primitive operations:
\emph{Collision detection} (CD), which is used to asses if a configuration is collision-free or not, and
\emph{Nearest neighbor} (NN) search, which is used to efficiently return the neighbor (or neighbors) of a given configuration.
The CD operation is also used to test if the straight line connecting two configurations lies in \Cfree---a procedure referred to as \emph{local planning}.
While in theory the cost of NN exceeds that of local planning,
in practice it is the latter that is the main computational bottleneck
in sampling-based \MPd algorithms~\cite{CBHKKLT05}.

However, recent results (e.g.~\cite{BKOF12, SH14-arxiv, LH14}) suggest that this may not always be the case.
By carefully replacing many expensive calls to the local planner with NN queries, one may reduce the running time of different sampling-based \MPd algorithms.
As a result, the computational overhead of NN queries plays a significant role in the running time of these algorithms.
Moreover, existing algorithms that ensure
\emph{asymptotic optimality}\footnote{An algorithm is said to be asymptotically optimal, if the cost of the solution produced by the algorithm asymptotically approaches the cost of the optimal solution. The notion of cost depends on the problem at hand and may be, e.g., path length, energy consumption along the path or minimal distance from the obstacles.}
such as PRM*~\cite{KF11}, FMT*~\cite{JP13} and MPLB~\cite{SH14-arxiv}
\textVersion{}{or
\emph{asymptotic near-optimality}\footnote{An algorithm is said to be asymptotically near-optimal, if given an approximation factor $\varepsilon$, the cost of the solution produced by the algorithm approaches a cost within a factor of $(1 + \varepsilon)$ of the optimal solution.}
(such as ANO-MPLB~\cite{SH14-arxiv})}
can make use of \emph{all-pairs $r$-nearest-neighbors} queries.
That is,
given a set $P$ of $n$ points and a radius $r = r(n)$
report all pairs of points $p,q \in P$ such that the distance between $p$ and $q$ is at most $r$.
This calls for application-specific NN data structures tailored to efficiently answering this type of queries.
Many implementations of efficient NN data structures exist, e.g., ANN~\cite{AMNSW98}, FLANN~\cite{flann}, and E2LSH~\cite{AI06}.
However, none of the above methods is tailored for these very specific NN queries that arise in the context of
these motion-planning algorithms.

\par
\noindent\textbf{Contribution and paper organization.}
This paper adopts Randomly Transformed Grids (RTG),
an algorithm by Aiger et al.~\cite{AKS13}
for finding all-pairs $r$-\nns, to sampling-based \MPd algorithms.
We begin by identifying which algorithms can make use of all-pairs~$r$-nearest-neighbors and review them in Section~\ref{sec:prelim}.
Specifically, we discuss the subtleties of using them in an anytime mode versus a batch mode.
After an overview of existing NN data structures
we present in Section~\ref{sec:rtg} the RTG algorithm together with
a description of our efficient implementation
(which will be publicly available, together with additional experimental results in our web-page\footnote{\url{http://acg.cs.tau.ac.il/projects/rtg}}).
We then proceed 
with
a series of experimental results comparing our implementation with
different state-of-the-art implementations of NN data structures and show the
speedup that is obtained by using our RTG implementation for all-pairs~$r$-\nns queries.
To test the affect of RTG in \MPd algorithms,
we present a series of simulations demonstrating that RTG allows to speed up the construction time of PRM-type roadmaps, the time to obtain an initial solution or to converge to high-quality solutions in complex scenarios.
For example, the construction time of PRM-type roadmaps in certain scenarios is reduced by a factor of between two and three.
We conclude in Section~\ref{sec:future} with a discussion of the current limitations of our approach together with suggestions for possible future work. 
\section{Preliminaries}
\label{sec:prelim}
We first review several sampling-based \MPd algorithms that 
can rely on 
NN queries of type all-pairs~$r$-nearest-neighbors.
We then continue to discuss existing NN data structures.
\subsection{Sampling-based motion-planning algorithms}
\textVersion{}
{
Throughout this subsection
we use the following procedures, which are standard procedures used in sampling-base \MPd algorithms.
\texttt{sample\_free}$(n)$ is a procedure returning $n$ random free configurations.
\texttt{nearest\_neighbor}$(x,V)$ and \texttt{$r$-nearest\_neighbors}$(x,V,r)$ return the nearest neighbor and all nearest neighbors in a ball of radius $r$ of $x$ within the set~$V$, respectively.
Let \texttt{steer}$(x,y)$ return a configuration $z$ that is closer to $y$ than $x$ is.
The procedure \texttt{collision\_free}$(x,y)$ tests whether the straight-line segment connecting $x$ and $y$ is contained in \Cfree, and
\texttt{dist}$(x,y)$  returns the Euclidean distance of the straight-line path connecting $x$ and $y$.
Let us denote by $\texttt{cost}_{\calG}(x)$ the minimal cost\footnote{In this paper, unless stated otherwise, we use Euclidean distance as the cost function.} of reaching a node $x$ from $x_{init}$ using a roadmap~\calG.
}

We begin by discussing the subtle difference between batch and anytime \MPd algorithms.
We refer to an algorithm as a \emph{batch} algorithm if it processes a predefined number of samples $n$ in one go.
An algorithm is said to be \emph{anytime} if it refines its solution as time progresses and may be run for any given amount of time.
Observe that efficiently computing all-pairs $r$-\nns
is intrinsically a \emph{batch} operation.
We briefly describe a scheme to easily modify a batch algorithm into an anytime one (see, e.g.,~\cite{WBC13}).
First, run the algorithm on an initial (small) set of $n$ samples.
Then, as long as time permits, double $n$ and re-run the algorithm using the larger $n$.
This way we obtain anytime algorithms,
which rely on all-pairs~$r$-nearest-neighbors---%
this in turn makes them amenable to the optimization that we propose in this paper.
We note that between iterations, additional optimizations such as pruning, informed sampling~\cite{GSB14}, using lower bounds~\cite{SH14-arxiv} or relaxing optimality~\cite{SH13} may be applied.

\textVersion{
Arguably, the best-known \MPd algorithm that makes use of all-pairs $r$-nearest-neighbors is PRM*~\cite{KF11}.
PRM* is a multi-query, batch, asymptotically-optimal algorithm that maintains a graph data structure as its roadmap.
It samples~$n$ collision-free configurations, which are the vertices of the roadmap.
Two configurations are connected by an edge if their distance is less than~$r(n)$
and if the straight line connecting them is collision-free. }
{Arguably, the best-known \MPd algorithm that makes use of all-pairs $r$-nearest-neighbors is PRM*~\cite{KF11}.
PRM*, outlined in Alg.~\ref{alg:prm_star}, is a multi-query, batch, asymptotically-optimal algorithm that maintains a graph data structure as its roadmap.
It samples $n$ collision-free configurations which are the vertices of the roadmap (line 1).
Two configurations are connected by an edge if their distance is less than $r(n)$
and if the straight-line connecting them is collision-free (lines 2-5).}
Specifically, the radius used~\cite{KF11} is
\begin{equation}
    r_{\text{{\tiny PRM}}^*} = 2
                \left[\left(1+ \frac{1}{d} \right)\cdot
                      \left( \frac{\mu(\Cfree)}{\zeta_d}\right)\cdot
                      \left( \frac{\log n}{n}\right)\right]^{1/d},
\label{eq:r_prm}				
\end{equation}
where
$d$ is the dimension,
$\mu(\Cfree)$ is the volume of the free space and
$\zeta_d$ is the volume of a $d$-dimensional sphere of radius~1.

\textVersion
{}
{
\begin{algorithm}[t,b]
\caption{PRM* (n)}
\label{alg:prm_star}
\begin{algorithmic}[1]
	\STATE	$V \leftarrow \set{x_{\text{init}}} \cup \texttt{sample\_free}(n)$;
					$E \leftarrow \emptyset$;
					$\calG\leftarrow (V,E)$
 \FORALL {$(x, y \in V)$}
 	\IF {(\texttt{dist}($x, y$) $ \leq r(n)$)}
 		\IF {(\texttt{collision\_free}($x, y$))}
	 		\STATE	$E \leftarrow E \cup (x,y)$
	 	\ENDIF
	\ENDIF
 \ENDFOR
 	
\end{algorithmic}
\end{algorithm}
}
To reduce the number of calls to the local planner,
one can delay local planning to the query phase and test if two neighbors are connected only if they potentially lie on the shortest path to the goal.
This lazy approach was originally suggested for the PRM algorithm~\cite{BK00}.
We apply this approach to the aforementioned batched PRM* and call it LazyB-PRM*.
Somewhat similarly, Luo and Hauser~\cite{LH14} have recently proposed an anytime,
\emph{single-query} variant for PRM* called Lazy-PRM*
that relies on dynamic shortest path algorithms to efficiently update the roadmap.

\textVersion
{
Janson and Pavone~\cite{JP13} introduced the Fast Marching Tree algorithm (FMT*).
The asymptotically-optimal algorithm can be viewed as a single query variant of PRM*.
Similarly to PRM*, FMT* samples $n$ collision-free nodes.
It then builds a minimum-cost spanning tree rooted at the initial configuration by growing in cost-to-come space.
The algorithm was shown to converge to an optimal solution faster than PRM* and RRT*~\cite{KF11}.
}
{
Janson and Pavone~\cite{JP13} introduced the Fast Marching Tree algorithm (FMT*).
The single-query asymptotically-optimal algorithm, outlined in Alg.~\ref{alg:fmt}, maintains a tree as its roadmap.
Similarly to PRM*, FMT* samples $n$ collision-free nodes $V$ (line 1).
It then
builds a minimum-cost spanning tree rooted at the initial configuration by maintaining two sets of nodes~$H, W$ such that
$H$ is the set of nodes added to the tree that may be expanded and $W$ is the set of nodes not in the tree yet~(line~2).
It then computes for each node the set of nearest neighbors\footnote{The nearest-neighbor computation can be delayed and performed only when needed but we present the batched mode of computation to simplify the exposition.
The delayed variant makes use of multiple $r$-nearest neighbors queries while the batched-mode variant makes use of all-pairs $r$-nearest neighbors queries.} of radius~$r(n)$~(line~4).
The algorithm repeats the following process: the node $z$ with the lowest cost-to-come value is chosen from $H$ (line 5 and 17).
For each neighbor~$x$ of~$z$ that is not already in~$H$, the algorithm finds its neighbor~$y \in H$ such that the cost-to-come of~$y$ added to the distance between~$y$ and~$x$ is minimal~(lines~8-10).
If the local path between~$y$ and $x$ is free,~$x$ is added to~$H$ with~$y$ as its parent~(lines~11-13).
At the end of each iteration~$z$ is removed from~$H$~(line~14).
The algorithm runs until a solution is found or there are no more nodes to process.
The algorithm, together with its bidirectional variant~\cite{SSJP14}, were shown to converge to an optimal solution faster than PRM* and RRT*~\cite{KF11}.
}
The radius used by the FMT* algorithm~\cite{JP13} is
\begin{equation}
	r_{\text{\tiny FMT}^*} = 2 (1 + \eta)
                    \left[\left(\frac{1}{d} \right)\cdot
                         \left( \frac{\mu(\Cfree)}{\zeta_d}\right)\cdot
                          \left( \frac{\log n}{n}\right)\right]^{1/d},
\label{eq:r_fmt}				
\end{equation}
where
$\eta>0$ is some small constant.
Moreover, Janson and Pavone show that PRM* can also use the (smaller) radius defined in Eq.~\ref{eq:r_fmt} while maintaining its asymptotic optimality.

\textVersion
{}
{
\begin{algorithm}[t, b]
\caption{FMT* $(x_{init}, n)$}
\label{alg:fmt}
\begin{algorithmic}[1]
	\STATE	$V \leftarrow \set{x_{\text{init}}} \cup \texttt{sample\_free}(n)$;
					$E \leftarrow \emptyset$;
					$\calT\leftarrow (V,E)$
  \STATE	$W \leftarrow V \setminus \set{x_{\text{init}}}$;
  				\hspace{3mm}
					$H \leftarrow \set{x_{\text{init}}}$
  \FORALL{$v \in V$}
  	\STATE	$N_v \leftarrow
  					\texttt{nearest\_neighbors}(V \setminus \set{v}, v, r(n))$
  \ENDFOR


  \STATE	$z \leftarrow x_{\text{init}}$
	\WHILE {$z \notin \calX_{\text{Goal}}	$}
		\STATE $H_{\text{new}} \leftarrow \emptyset$;
					 \hspace{3mm}
					 $X_{\text{near}} \leftarrow W \cap N_z$

		\FOR  {$x \in X_{\text{near}}$}
			\STATE $Y_{\text{near}} \leftarrow H \cap N_x$
							\hspace{3mm}
			\STATE $y_{\text{min}} \leftarrow \arg \min_{y \in Y_{\text{near}}}
								\set{\texttt{cost}_{\calT}(y) + \texttt{dist}(y,x)} $

			\IF {\texttt{collision\_free}$(y_{\text{min}}, x)$}
				\STATE $\calT.\texttt{parent}(x) \leftarrow y_{\text{min}}$
				\STATE $H_{\text{new}} \leftarrow H_{\text{new}} \cup \set{x}$;
							 \hspace{3mm}
					 		 $W \leftarrow W \setminus \set{x}$
			\ENDIF
		\ENDFOR
								
					
		\STATE $H \leftarrow (H \cup H_{\text{new}}) \setminus \set{z}$
		\IF {$H = \emptyset$}
			\RETURN FAILURE
		\ENDIF
		
		\STATE $z \leftarrow \arg \min_{y \in H}
								\set{\texttt{cost$_{\calT}$}(y)} $
														
	\ENDWHILE

	\RETURN PATH \hspace{3mm}
\end{algorithmic}
\end{algorithm}						
}

Recently, we proposed a scheme to compute tight, effective lower bounds on the cost to reach the goal~\cite{SH14-arxiv}.
Incorporating these bounds with the FMT* algorithm, we introduced Motion Planning using Lower Bounds or MPLB.
The algorithmic tools used by MPLB cause the weight of collision-detection to be negligible when compared to NN calls.
Some of the experimental results suggest that more than 40\% of the running time of the algorithm is spent on NN queries.
MPLB uses the same radius as FMT* (Eq.~\ref{eq:r_fmt}).

Both FMT* and MPLB perform $r$-nearest-neighbors
queries for a subset of the input points (we call this \emph{multiple $r$-\nns queries)}.
On the other hand, the NN data structure we propose to use answers only all-pairs $r$-nearest-neighbors queries.
This can easily be addressed by performing an all-pairs $r$-\nns query once, and storing all such pairs.
Multiple $r$-nearest-neighbors queries are then reduced to
querying the stored set of pairs and returning the relevant ones.
Clearly, this will only be efficient when the number of multiple $r$-nearest-neighbors queries is large.
As demonstrated in Section~\ref{sec:exp} this is indeed the case.

\textVersion
{
In the extended version of our paper~\cite{KSH14-arxiv} we present another
algorithm that makes use of multiple $r$-nearest-neighbor queries in addition to nearest-neighbor ones.
The algorithm, which we call batched-RRT*, is a
variant of the single-query
asymptotically-optimal RRT* algorithm~\cite{KF11}.
%
}
{
RRT*~\cite{KF11} is a single-query asymptotically-optimal variant of the RRT algorithm~\cite{KL00}.
We first outline the RRT algorithm (lines 1-7 of Alg.~\ref{alg:rrt_star}) and then continue
describing the RRT* algorithm together with a variant which we call batched-RRT*.
The RRT algorithm maintains a tree as its roadmap.
At each iteration a configuration $x_{rand}$ is sampled at random (line~3).
Then, $x_{nearest}$, the nearest configuration to $x_{rand}$ in the roadmap is found (line~4) and extended in the direction of $x_{rand}$ to a new configuration~$x_{new}$ (line~5).
If the path between $x_{nearest}$ and $x_{new}$ is collision-free, then $x_{new}$ is added to the roadmap (lines~6-7).

\begin{algorithm}[t,b]
\caption{RRT* ($x_{init}, n$ )}
\label{alg:rrt_star}
\begin{algorithmic}[1]
	\STATE	$V \leftarrow \set{x_{\text{init}}}$;
					\hspace{1mm}
					$E \leftarrow \emptyset$;
					\hspace{1mm}
					$\calT\leftarrow (V,E)$
	\FOR  {$i = 1 \ldots n$}
	 	\STATE $x_{rand} \leftarrow \texttt{sample\_free()}$
 		\STATE $x_{nearest} \leftarrow
 												\texttt{nearest\_neighbor}(	 x_{rand}, V)$
 		\STATE $x_{new} \leftarrow
 												\texttt{steer}(	x_{nearest}, x_{rand})$

 		\IF {(\texttt{collision\_free}($x_{nearest}, x_{new}$))}
 			\STATE	$V \leftarrow V \cup \set{x_{new}}$;
							\hspace{1mm}
 							$\calT.\texttt{parent}(x_{new}) \leftarrow x_{nearest}$
 			\vspace{3mm}
 			  \STATE $X_{near} \leftarrow
 			  					r\texttt{-nearest\_neighbors}(x_{new}, V, n)$
 			 	\FORALL {$(x_{near}, X_{near})$}
 			 		\STATE  \texttt{rewire\_RRT$^*$}($x_{near}, x_{new}$ )
  			\ENDFOR
  			\FORALL {$(x_{near}, X_{near})$}
			  	\STATE  \texttt{rewire\_RRT$^*$}($x_{new}, x_{near}$ )
			  \ENDFOR
		\ENDIF
	\ENDFOR
\end{algorithmic}
\end{algorithm}

\begin{algorithm}[t, b]
\caption{\texttt{rewire\_RRT$^*$}($x_{potential\_parent}, x_{child}$)}
\label{alg:rewire}
\begin{algorithmic}[1]

  \IF {(\texttt{collision\_free}($x_{potential\_parent}, x_{child}$))}
  	\STATE	$c \leftarrow$ \texttt{dist}($x_{potential\_parent}, x_{child}$)
  	\IF {	($\texttt{cost}_{\calT}(x_{potential\_parent}) + c
  						<
  				  \texttt{cost}_{\calT}(x_{child}))$}
	    \STATE	$\calT.\texttt{parent}(x_{child}) \leftarrow
    						x_{potential\_parent}$
    \ENDIF
 	\ENDIF

\end{algorithmic}
\end{algorithm}						

RRT* follows the same steps as the RRT algorithm but has an additional stage after an edge is added to the tree (lines 8-12 of Alg.~\ref{alg:rrt_star}):
a set $X_{near}$ of the $r$-nearest neighbors of $x_{new}$ is considered and a \emph{rewiring} step (see Alg.~\ref{alg:rewire}) occurs twice:
first, it is used to find the node $x_{near} \in X_{near}$ which will minimize the cost to reach $x_{new}$;
then, the procedure is used to attempt to minimize the cost to reach every node $x_{near} \in X_{near}$ by considering $x_{new}$ as its parent.
We note that RRT* uses two types of NN queries:
both nearest-neighbor and multiple $r$-nearest\_neighbors.
Moreover, the original formulation of RRT*~\cite{KF11} uses at step $i$ a radius of $r_{RRT*}(i)$ where
\begin{equation}
	r_{\text{\tiny RRT}^*}(i) =	\left[\left( 2\left(1 + \frac{1}{d}\right) \right)\cdot
                        \left( \frac{\mu(\Cfree)}{\zeta_d}\right) \cdot
                        \left( \frac{\log i}{i}\right)\right]^{1/d}.
\label{eq:r_rrt}				
\end{equation}
However, to ensure asymptotic optimality,
a radius of $r(n)$ may be used at each stage of the algorithm
(see proof of Thm.~38 in~\cite{KF11}).

We propose a batch variant of RRT* that stems from the FMT* framework.
Instead of using $n$ uniformly sampled configurations,
one may use the $n$ nodes constructed by an RRT algorithm and build the FMT*-tree using these nodes.
This variant benefits from
(i)~the fast exploration of the configuration space due to the Voronoi-bias that RRT has and
(ii)~the efficient construction of the shortest-path spanning tree due to the Dijkstra-like pass that FMT* has.
We call this variant batched-RRT*.
}

We summarize the different algorithms in Table~\ref{tbl:algs} together with the type of NN queries that they use.

\begin{table}
	\centering
	\caption{\sf	\footnotesize
  					List of algorithms that use $r$-nearest neighbors queries.}
  \label{tbl:algs}
	\begin{tabular}{| l || l | l |}
    \hline
 		Algorithm					& NN queries used		& Comments	\\
 		\hline \hline
 		PRM*							& all-pairs $r$-NN  	& Multi-query					 \\ \hline
		LazyB-PRM*	&	all-pairs $r$-NN		&	Multi-query					 \\ \hline
											& 			  						 & All-pairs variant exists \\
 		FMT*							& multiple $r$-NN  		& Single-query 				\\ \hline
											& 			  						 & All-pairs variant exists \\ 				
MPLB							& multiple $r$-NN  	& Single-query, anytime	 \\ \hline
 											& nearest neighbor					 & All-pairs variant exists\\
 		batched-RRT*			&	multiple $r$-NN  	&
 			\textVersion{Single-query, anytime, see~\cite{KSH14-arxiv}}{}		 \\ \hline
  \end{tabular}
\end{table}

\subsection{Review of existing nearest-neighbor data structures}
As nearest-neighbors search is widely used in various domains
there exists a wide range of methods allowing efficient proximity queries,
differing in their space requirement and
time complexity.
Such methods include the
\kdtrees~\cite{B75,FBF77},
geometric near-neighbor access trees (GNAT)~\cite{B95},
locality sensitive hashing (LSH)~\cite{IM98}, and others~\cite{GG98}.

\kdtrees, which work best for rather low dimensions, are often used in motion-planning settings.
A \kdtree is a binary tree storing the input points in its leaves,
where each node~$v$ defines an axis-aligned hyper-rectangle
containing the points stored in the subtree rooted at~$v$.
Given a query point~$q$, NN search is performed in two phases:
the first locates the leaf node with the hyper-rectangle containing~$q$,
and the second traverses the tree backwards searching for closer sibling nodes.
Given a~$d$-dimensional point set of size~$n$, construction takes $O(dn\log{n})$ time.
Friedman et al.~\cite{FBF77} showed that under mild assumptions
the expected time for a single \nn query is $O(\log{n})$.
For $r$-\nns queries,
the expected complexity is at least $\Omega(\log{n} + k)$, where $k$ is the number of reported neighbors
(the worst-case bounds are much worse~\cite{BCKO08}).

Another recursive structure that is frequently used in
motion-planning algorithms is GNAT.
The input point set is recursively divided into smaller subsets and
each subset is then represented using a subtree.
Searching the structure is done recursively,
while the recursion call continues to child nodes that have not yet been pruned.
As claimed in~\cite{B95}, typically only linear space is required
and the construction takes $O(dn\log{n})$ time.

A common practice for speeding up algorithms that use NN queries
but do not require exact results (such as \MPd algorithms)
is to use \emph{approximate} NN queries.
Different types of approximate $r$-\nns methods exists:
some (e.g.,~\cite{AMNSW98}) return with high probability most neighbors of a given query point~$q$,
while others
return a neighbor within a distance~$r(1+\epsilon)$ if
a neighbor
at distance of at most~$r$ from~$q$ exists.


Locally sensitive hashing (LSH), presented by Indyk and Motwani~\cite{IM98},
is an approximate \nn method
for~$d$-dimensional point sets.
As opposed to many other methods, its $O(dn^{\frac{1}{1+\epsilon}})$ query time does
not depend exponentially on the dimension.
LSH uses a subset of~$t$ hash functions from a family of locally sensitive hash functions for mapping the data into buckets.
That is,
with high probability two close points will be mapped to the same bucket.
Given a query point~$q$, the algorithm maps~$q$ using the set of hash functions to a set of buckets and
collects all data points that were mapped to these buckets.
The required neighbor is then found within this set of collected candidates.

\textVersion{}
{
An algorithm for finding all nearest-neighbors in a given fixed $d$-dimensional point set
under the $L_\infty$-metric was originally
presented by Lenhof and Smid~\cite{LS95}.
The algorithm uses a fixed uniform grid for inspecting
pairs of points that lie in the same grid cell or in two adjacent ones.
Its time complexity is linear in both the input size and the output size.
A simplified variant was later presented by Chan~\cite{C01}.
}

\section{Randomly Transformed Grids (RTG)}
\label{sec:rtg}

Aiger et al.~\cite{AKS13} suggest two simple
randomized algorithms for approximately answering
all-pairs $r$-\nns queries
given a set $P$ of $n$ points in a $d$-dimensional Euclidean space.
\textVersion{
(See~\cite{AKS13} for earlier results that rely on grids for batch NN queries in other metric spaces.)}
{}

The first algorithm, 
outlined in Alg.~\ref{alg:rsg},
conceptually places a $d$-dimensional axis-parallel grid of cell size\footnote{
\label{footnote:c_tilde_def}
The cell size $c$ should be slightly larger than the radius $r$.
Thus, to specify the cell size,
one needs to specify a constant, which we call the cell-size factor,
$\tilde{c}(r, d) > 1$,
such that $c = \tilde{c} \cdot r$.
}~$c$,
which is shifted
according to a randomly chosen uniform shift (line~2).
This grid defines a partition of the points in $P$ into cells,
and each point is associated to the cell $u$ containing it (lines~4-7).
For each non-empty grid cell, the distance between every two points
associated to the cell is computed. A pair inside a cell is then reported if the
computed distance is at most~$r$ (lines~8-9).
The process is repeated~$m$ times to
guarantee
that, with high probability, most pairs of points at Euclidean distance at most $r$ will be reported.
The second algorithm follows the same
approach adding a random orientation to each randomly shifted grid.

\begin{algorithm}[t,b]
\caption{Randomly Shifted Grids ($P,r,c,m$ )}
\label{alg:rsg}
\begin{algorithmic}[1]

 \FOR {$i \in \set{1 \ldots m} $ }
 	\STATE Choose a random shift for a grid of cell size $c$
    \STATE $U \leftarrow \emptyset$
 	\FORALL {$p \in P$}
        \STATE Compute the grid cell $u$ that $p$ lies in
        \STATE Associate $p$ to $u$
        \STATE $U \leftarrow U \cup \{u\}$
 	\ENDFOR
    \FORALL {$u\in U$}
        \STATE  Go over all pairs of points in $u$ and report those of
 					 Euclidean distance at most $r$
    \ENDFOR
 \ENDFOR
\end{algorithmic}
\end{algorithm}

Assuming a constant-cost
implementation of the floor function and of hashing,
Aiger et al.~\cite{AKS13} show that for appropriate choices of~$c$ and~$m$, with high probability,
the algorithm reports every pair at
distance at most $r$ with time $O((n+k) \log n)$,
where $k$ is the number of pairs at distance at most $r$.
Note that the hidden constant depends exponentially on the dimension $d$.
When only randomly shifted grids are used,
the constant is roughly $(.484 \sqrt{d})^d$.
However, when both rotations and translations are used,
it is roughly $6.74^d$.
For input sets lying in a
subset of the space that has low doubling dimension,
much tighter bounds can be obtained.

We next
discuss the effect of the algorithm's parameters, 
outline several tradeoffs,
and
describe an efficient implementation of the algorithm.

\subsection{The key parameters}
\rtg requires the user to set the cell size~$c$ and the number~$m$ of randomly shifted grids.
These two parameters have a major impact on the performance of the algorithm.
When a very small~$c$ or~$m$ is used,
the set of reported pairs might be small
compared to the true number of neighboring pairs.
However, when a large value of~$c$ is used,
the algorithm  may output the whole set of true neighbors
at the cost of performing
several inefficient brute-force searches.
Obviously, the more iterations performed (larger $m$) the better the results are.
Therefore, a tradeoff between the running time and the quality of the results exists,
and any subtle change to either~$c$ or~$m$ may affect both measures.
Experiments supporting this observation are detailed in Subsec.~\ref{subsec:exp:param_tune}.

\subsection{Implementation details}

Implementing the algorithm in a \naive manner is straightforward.
One has to store the non-empty cells and their associated points.
All pairs of points within a cell are examined in a brute force manner,
computing the distance of a pair
in $O(d)$ time and reporting the pair if relevant.

Note that a certain pair of points may be associated to the same cell
in several different grids.
Thus, in order to report every neighboring pair only once,
an auxiliary data structure for
storing the already reported pairs is needed.
If such a structure allows
efficient query operations as well as
efficient insertions, it
can be utilized
to avoid costly distance computations
by filtering out many potential pairs.
As a result, the running time of the algorithm can be significantly reduced
at the cost of storing the auxiliary pairs structure.

Let $P = \{p_1, \ldots ,p_n\}$ be the set of input points.
A possible auxiliary structure would be an array
of unordered sets, where the~$i$th cell of the array
stores all indices $j > i$ such that~$(p_i,p_j)$ is a reported pair.
The space complexity of such a structure is linear in the size of the output (the number of reported pairs),
whereas the expected query time and update time are constant.
On the other hand, one can use a different auxiliary structure,
whose space complexity is $O(n^2)$, and experience much better query and update times.
For instance,
a bit array representing a two-dimensional matrix, where the cell $(i,j)$ is set to one if the pair $(p_i, p_j)$ is a reported pair,
is a possible structure (notice that only half of the array needs to be stored due to symmetry).
It supports
constant-time insert and constant-time query operations.
Nevertheless, both the space and the query time complexity
do not depend on the output size.
However, such a solution is restricted to a limited number of input points,
depending on the machine on which the query is executed.
We refer to this structure as a flattened two-dimensional bit array.

Our \Cpp implementation 
supports either auxiliary data structures discussed.
The array of unordered sets is implemented as a vector of \texttt{boost::unordered\_set}-s, while \texttt{boost::dynamic\_bitset}
is used for implementing the flattened bit-array~\cite{S11}.
Although the latter typically exhibits running times that are twice as fast as the former (for dimensions six and above), it is highly non-scalable with respect to memory consumption, thus only applicable to settings where a limited number of samples is required.
In the rest of the paper we report on results of the first variant only,
namely an array of unordered sets.

As we construct one grid at a time,
and since each new grid may introduce new neighboring pairs,
the pairs are reported in an unordered manner.
Therefore, only all-pairs $r$-\nns queries are supported.
Yet, the structures can be extended such that they
support multiple $r$-\nns as~well.
In order to do so, 
the auxiliary structure should store every pair twice.
Then, after finding all pairs, multiple $r$-\nns queries can be easily answered.

We have also implemented the second algorithm proposed by Aiger et al. that
adds random orientations to the constructed grids.
Our implementation, which uses the Eigen \Cpp library~\cite{eigenweb},
did not achieve significant improvement in running time and quality of the results,
comparing to the first \rtg algorithm.
Therefore, we do not report on these results here.
We leave it for further research to understand the gap
between the optimistic theoretical prediction and the effect
of orientation in practice.

\section{Experimental results}
\label{sec:exp}
To evaluate our implementation,
we first report on a set of experiments aiming both to
demonstrate the sensitivity of the RTG algorithm to the parameters used and to determine the better (ideally, optimal) ones.
Using these computed parameters, we first compare our implementation with several
state-of-the-art implementations of NN data structures and show the speedup that may be obtained by using our RTG implementation for all-pairs $r$-\nns queries.
Finally, to test the effect of RTG in \MPd algorithms,
we present a series of simulations that demonstrate that RTG allows to speed up the construction time of a PRM-type roadmap, the time to obtain an initial solution or to converge to high-quality solutions.
All experiments were executed on a 2.8GHz Intel Core i7 processor with 8GB of RAM.
\textVersion{
We note that due to space limitation we present only a subset of our results.
For additional experiments and plots we refer the reader to the extended version of our paper~\cite{KSH14-arxiv}.
}

\subsection{Parameter tuning}
\label{subsec:exp:param_tune}
Recall that given a set of $n$ points in a $d$-dimensional
space and a radius $r=r(n)$ (as in Eq.~\ref{eq:r_fmt}),
the \rtg algorithm has two parameters that should be set:
the cell size $c$ and the number~$m$ of grids to construct.
As discussed in Sec.~\ref{sec:rtg}, the algorithm is rather sensitive with respect to either parameter.
%
%
Therefore,
we conducted the following experiments
in the unit~$d$-dimensional hypercube:
we executed our \rtg implementation
using increasing values of~$c$ and~$m$ and
measured both the time for running
all-pairs $r$-\nn and the success-rate, that is, the ratio
between the number of reported pairs and the ground truth.
We repeated the experiment for different values of $n$ and $d$.

\begin{figure}[]
  \centering
    \subfloat
   []
   {
   	\includegraphics[width=6.7cm]{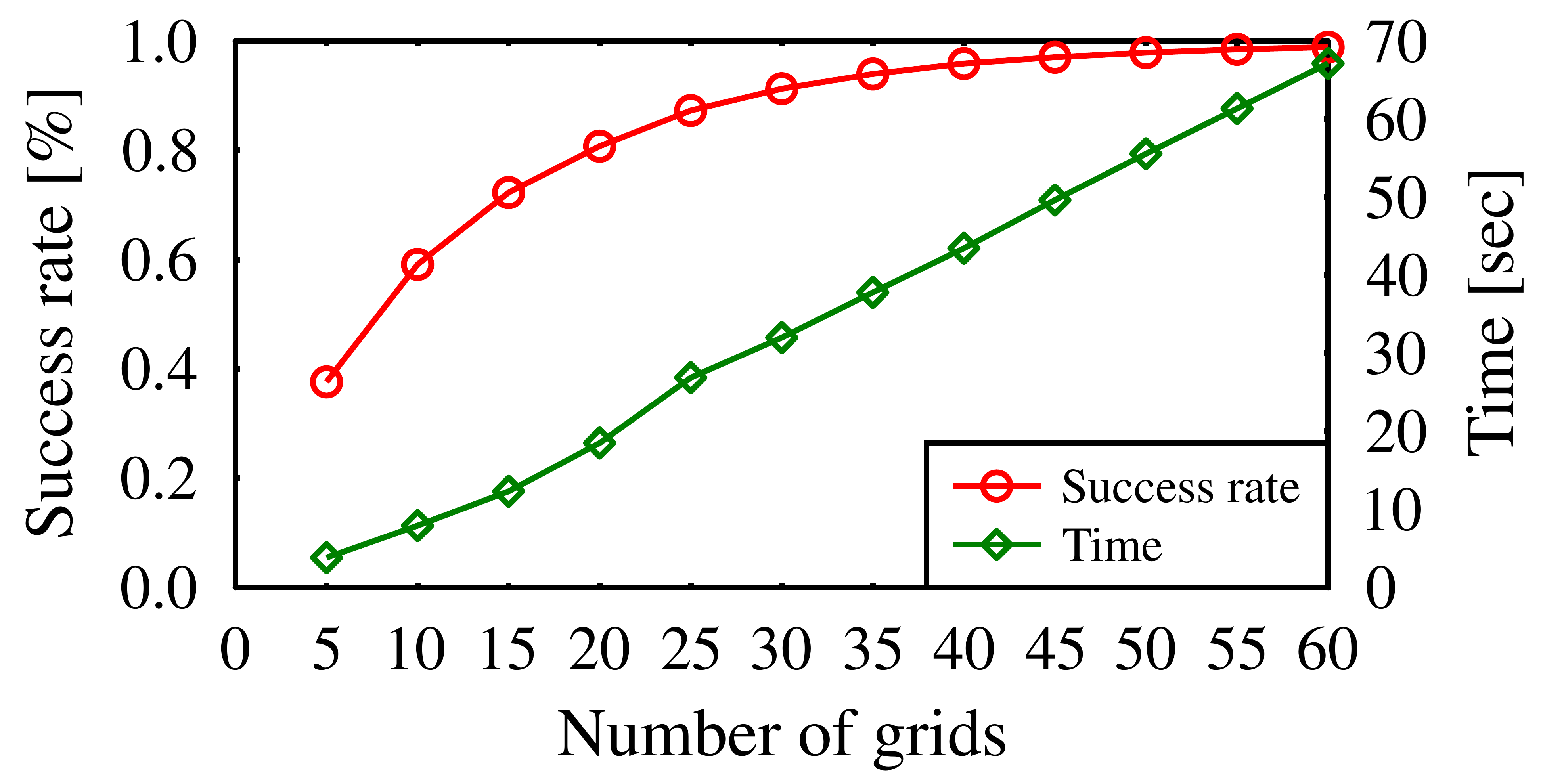}
   	\label{fig:t_and_sr_vs_m}
   }
   \\
   \vspace{-10pt}
   \subfloat
   []
   {
   	\includegraphics[width=6.7cm]{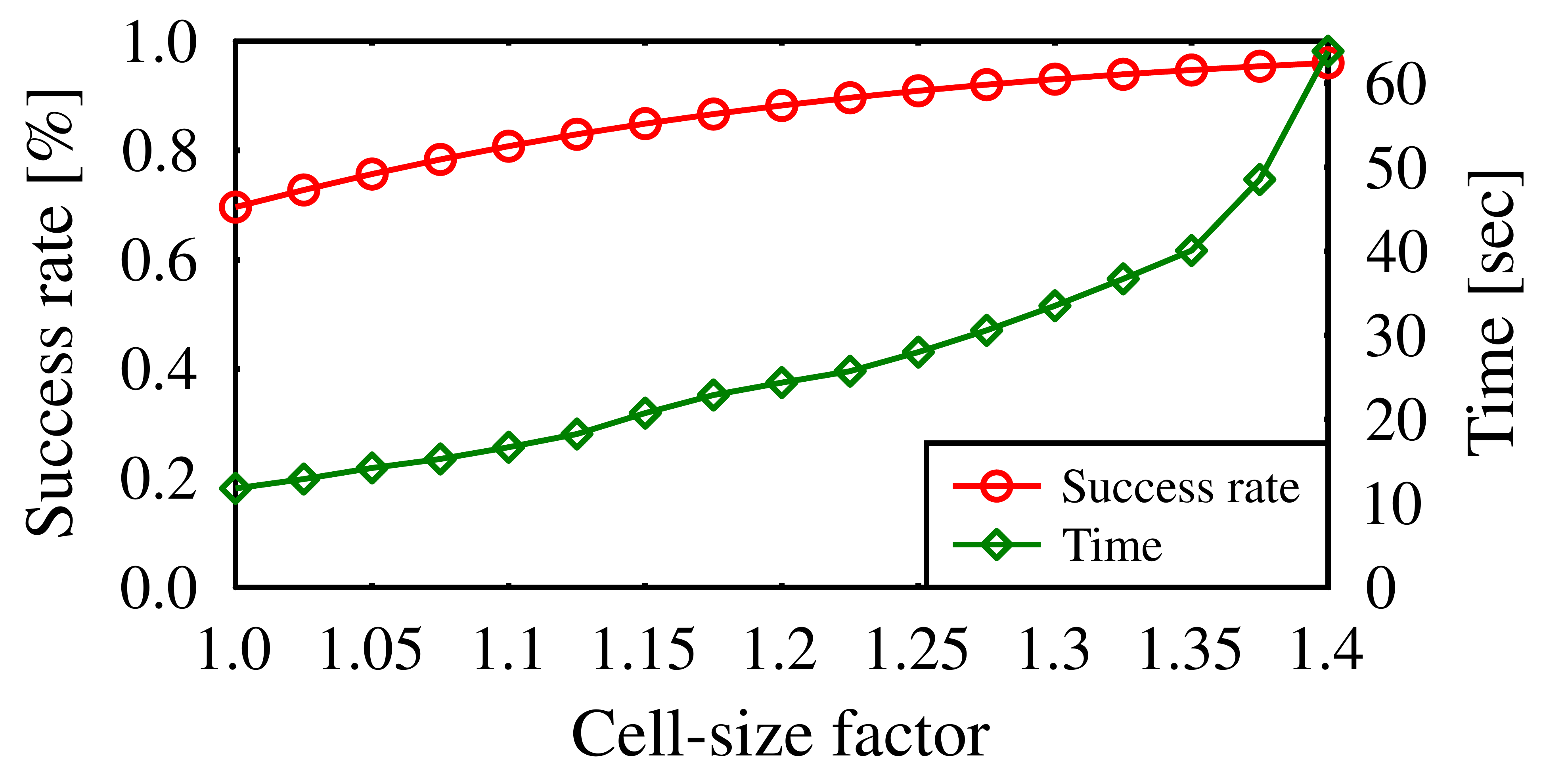}
   	\label{fig:t_and_sr_vs_c}
   }
   \vspace{-3pt}

  \caption[The effect of $c$ and $m$]
  { \sf Success rate (red, left axis) and  running time (green, right axis) as a function of
    (a)~the number $m$ of grids (for a fixed value of $\tilde{c}$ = 1.1)
    and
    (b)~the cell-size factor~$\tilde{c}$ (for a fixed number of grids $m=20$).
    Results are for~$n=102400$ and $d = 9$. }
  \label{fig:T_and_SR_vs_c_and_m}
\end{figure}
Fig.~\ref{fig:t_and_sr_vs_m} shows that
for a given cell-size factor $\tilde{c}$ (recall that $\tilde{c}= \frac{c}{r}$), increasing the number $m$ of grids results in
both an increase in the success rate of the algorithm as well as a linear growth in the running time.
Similar behavior was observed for a given $m$, when
$\tilde{c}$
gradually increases (Fig.~\ref{fig:t_and_sr_vs_c}).

Requiring a success rate of at least 98\%,
we chose for each~$d$ and~$n$ the values of $m$ and $\tilde{c}$
(and thus the value of~$c$)
that yielded the best running times.
\textVersion
{
For a table summarizing our results, see~\cite{KSH14-arxiv}.
}
{
Our results are summarized in Table~\ref{tbl:best_c_and_m}.
We note that similar behavior and results were obtained when running the same experiment on a point set
sampled in an environment cluttered with obstacles.
}

Throughout the next set of experiments, the values for $c$ and $m$ are selected according to the aforementioned table.
Note that Aiger et al.~\cite{AKS13}
use completely different values for~$c$ and~$m$,
which we found too crude for our setting.
The difference lies in the smaller radius
used in their experiments,
causing the output size to be very small with respect to $n$.
This is not surprising as their data comes from an application
of completely different nature.

\textVersion
{
}
{
\begin{table}
	\centering
	\caption{\sf	
  Input parameters ($m$ and $\tilde{c}$)
    for which the implementation obtained the fastest running times, while
    reporting at least 98\% of neighbors, as a function of the
    number of points~$n$ and the dimension~$d$.
  }
    \begin{tabular}{| l || l | l | l | l | l | l | l | l |}
    \hline
 		$d$		      & \multicolumn{2}{|c|}{3}     & \multicolumn{2}{|c|}{6}           & \multicolumn{2}{|c|}{9}          & \multicolumn{2}{|c|}{12}      \\
 		\hline
 	      $n$		      & $m$	  & $\tilde{c}$    & $m$     & $\tilde{c}$     & $m$     & $\tilde{c}$     & $m$    & $\tilde{c}$ \\
 		\hline
 		100	         &	  20  & 1.2	         &  20    & 1.35  	 &  20    & 1.4  & 30   & 1.4  \\\hline
 		200          &   20   & 1.275        &  20    & 1.35    &  20    & 1.4  & 30   & 1.4  \\ \hline
 		400          &  20    & 1.2          &  20    & 1.35    &  30    & 1.3  & 30   & 1.4 \\ \hline
 		800          &  20    & 1.2          &  25    & 1.25    &  25    & 1.325  & 30   & 1.4 \\ \hline
 		1600         &	 20   & 1.2          &  25    & 1.225   &  25    & 1.35  & 30   & 1.4 \\ \hline
        3200        &  20    & 1.15         &  20    & 1.35    &  25    & 1.35  & 30   & 1.325 \\ \hline
        6400        &  20    & 1.175        &  25    & 1.225   &  30    & 1.275  & 30   & 1.325 \\ \hline
        12800       &  20    & 1.15         &  20    & 1.35    &  35    & 1.225  & 55   & 1.175 \\ \hline
        25600       &  20    & 1.15         &  20    & 1.325   &  35    & 1.225  & 35   & 1.35 \\ \hline
        51200       &  20    & 1.15         &  20    & 1.35     &  40    & 1.2  & 30   & 1.425 \\ \hline
        102400      &  20    & 1.15         &  20    & 1.325    &  40    & 1.2  & N/A   & N/A \\ \hline
        204800      &  20    & 1.15         &  20    & 1.325   &  N/A   & N/A   & N/A   & N/A \\ \hline
  \end{tabular}

  \label{tbl:best_c_and_m}
\end{table}
}

\subsection{Comparison with existing NN libraries}

We compared our \rtg implementation with the following state-of-the-art NN methods:
FLANN \kdtree~\cite{flann}, ANN \kdtree~\cite{AMNSW98},
and LSH
in Euclidean metric spaces (E2LSH)~\cite{AI06}. %
\textVersion
{%
We invested particular efforts in tuning the parameters of all methods so as to obtain the best possible results
in each~\cite{KSH14-arxiv}.
}
{
For ANN
we set the bucket size to $\log{n}$, where $n$ is the number of input points,
and allowed an error bound of $\varepsilon=0.5$ in the query phase.
We mark that this had no major effect on the number of reported pairs.
For consistency reasons, we used the same error bound in FLANN.
For E2LSH we had to empirically estimate the optimal parameters for our point set.
Therefore, we used a training set and selected the parameters that minimize the running time on the training set while still
reporting a true neighbor with probability at least~0.9.
As mentioned,
for RTG we chose~$c$ and~$m$ according to Table~\ref{tbl:best_c_and_m}.
}
For each method we measured the time for answering an
all-pairs $r$-\nns queries\footnote{Recall that for some of the methods
finding all pairs requires $n$ single $r$-\nns calls,
with each of the $n$ points as an input.}
for~$n$ random uniform samples from
the unit~$d$-dimensional hypercube.
The radius~$r=r(n)$ was defined as in Eq.~\ref{eq:r_fmt}.
We used point sets, of increasing sizes, of dimensions~$d=3,6,9,\text{ and }12$.

\textVersion{
The results for nine dimensions, averaged over ten different runs, are presented in Fig.~\ref{fig:NN_compare}.
Clearly, as the number of samples increases
(exactly where NN dominates the running times of MP algorithms)
the gap in running time between our RTG implementation and the other methods grows.
}
{
The results for different dimensions,
averaged over ten different runs, are presented in Fig.~\ref{fig:NN_compare_3D},~\ref{fig:NN_compare_9D} and~\ref{fig:NN_compare_12D}.
Clearly, as the number of samples increases
(exactly where NN dominates the running times of MP algorithms)
the gap in running time between our RTG implementation and the other methods grows.
Similar results (see Fig.~\ref{fig:NN_compare_6D_cubicles}) were obtained when the environment was cluttered with obstacles.
}

\textVersion
{
\begin{figure}[t,b]
  \centering
    \includegraphics[width=6.7cm]{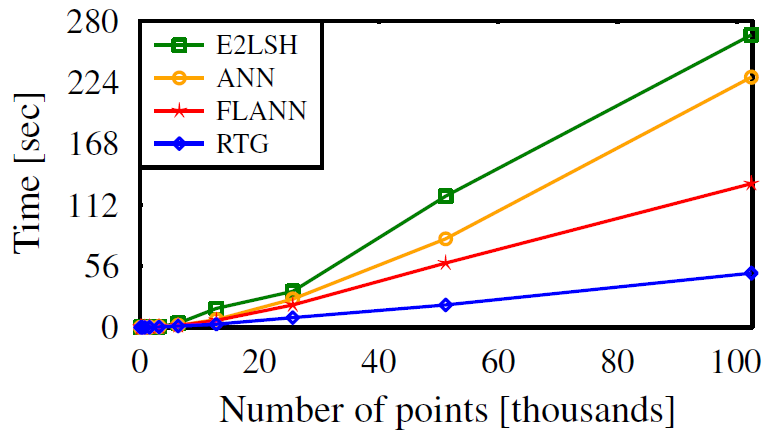}
 \vspace{-7pt}
  \caption{ \sf \footnotesize
  					Comparison between NN methods running all-pairs $r$-\nns for
  					randomly sampled points in the nine-dimensional unit hyper-cube.}

  \label{fig:NN_compare}
\end{figure}
}
{
\begin{figure}[t,b]
  \centering
    \includegraphics[width=6.7cm]{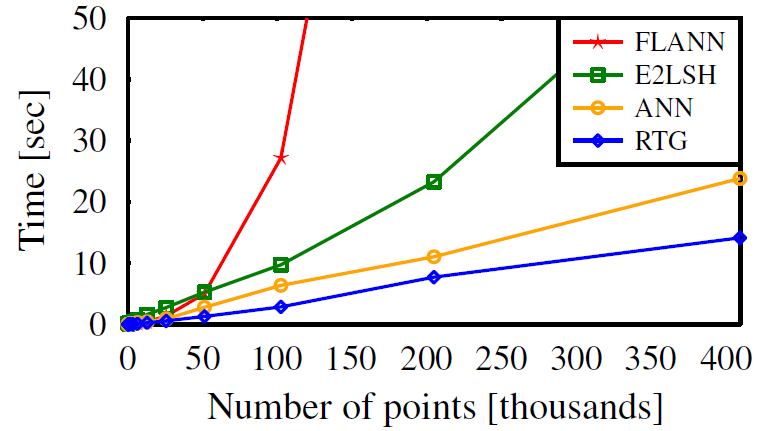}
  \caption{ \sf \footnotesize
  					Comparison between NN methods running all-pairs $r$-\nns for
  					randomly sampled points in the three-dimensional unit hyper-cube.}
  \label{fig:NN_compare_3D}
\end{figure}
\begin{figure}[t,b]
  \centering
    \includegraphics[width=6.7cm]{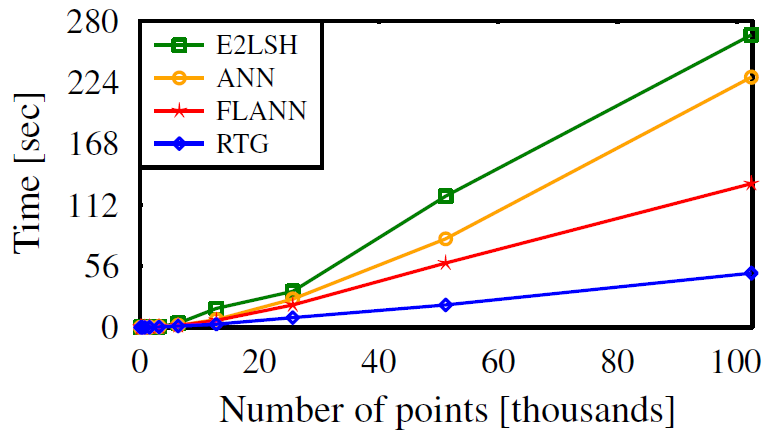}
  \caption{ \sf \footnotesize
  					Comparison between NN methods running all-pairs $r$-\nns for
  					randomly sampled points in the nine-dimensional unit hyper-cube.}
  \label{fig:NN_compare_9D}
\end{figure}
\begin{figure}[t,b]
  \centering
    \includegraphics[width=6.7cm]{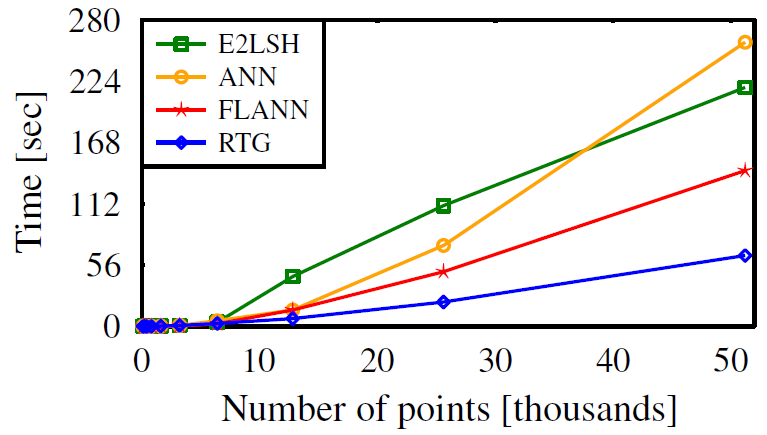}
  \caption{ \sf \footnotesize
  					Comparison between NN methods running all-pairs $r$-\nns for
  					randomly sampled points in the twelve-dimensional unit hyper-cube.}
  \label{fig:NN_compare_12D}
\end{figure}
\begin{figure}[t,b]
  \centering
    \includegraphics[width=6.7cm]{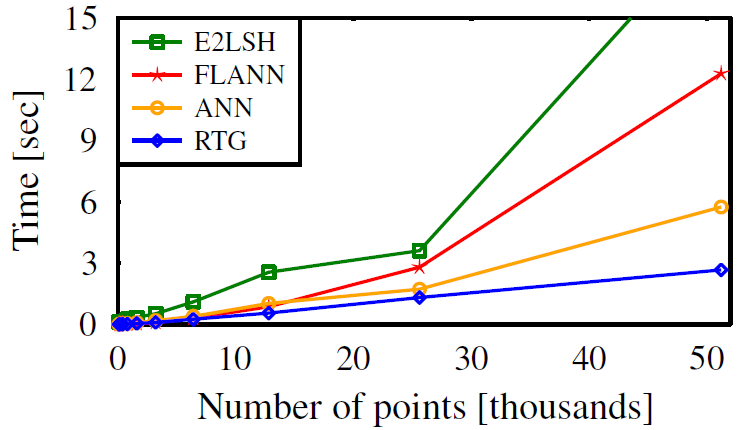}
  \caption{ \sf \footnotesize
  					Comparison between NN methods running all-pairs $r$-\nns for
  					randomly sampled points in an environment cluttered with obstacles (see the Cubicles scenarios, Fig.~\ref{fig:cubicles}).}
  \label{fig:NN_compare_6D_cubicles}
\end{figure}
}

\subsection{RTG in motion-planning algorithms}
\begin{figure}[t,b]
  \centering
  \subfloat
   [\sf Z-tunnel]
   {
   	\includegraphics[width = 0.3\textwidth ]{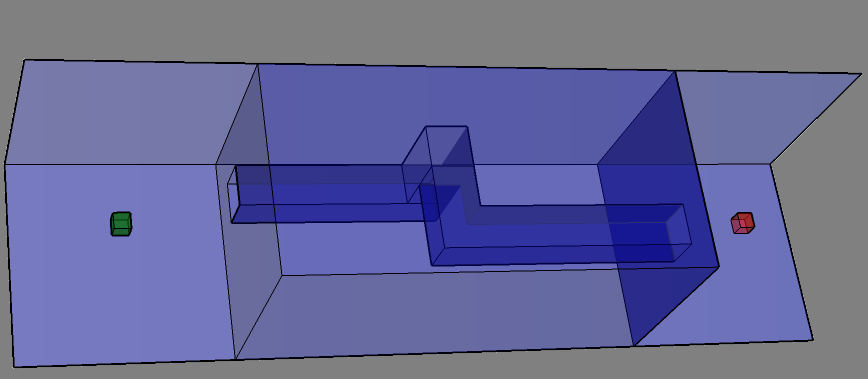}
   	\label{fig:corr}
   }
   \\
   \vspace{-5pt}
  \subfloat
   [\sf 3D Grid]
   {
   	\includegraphics[width = 0.28\textwidth]{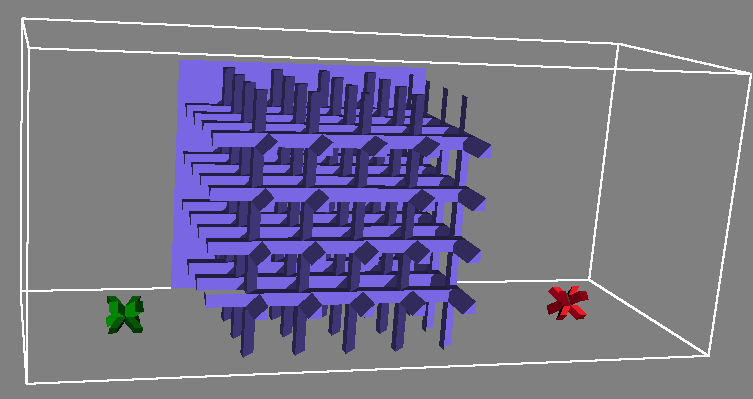}
   	\label{fig:3dgrid}
   }
   \subfloat
   [\sf Cubicles]
   {
   	\includegraphics[width = 0.159\textwidth]{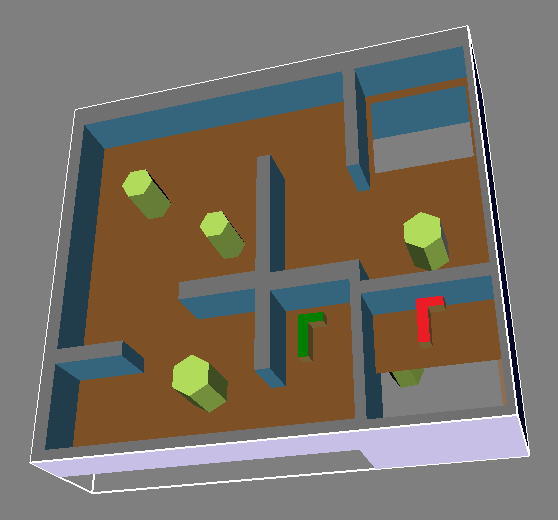}
   	\label{fig:cubicles}
   }
  \caption{\sf	\footnotesize
    Scenarios used for \MPd experiments.
    When testing the single-query algorithms
    the green and red robots need to interchange positions.
    						}
  \label{fig:scenarios}
\end{figure}
We integrated our RTG implementation within the OMPL~\cite{SMK12} framework, which uses GNAT as its primary NN structure.
This allowed us to
compare the two NN data structures in different \MPd algorithms.
The scenarios we used are depicted in Fig.~\ref{fig:scenarios}.
Each result is averaged over 50 runs.

\begin{figure}[t]
  \centering
   \includegraphics[width=6.7cm]{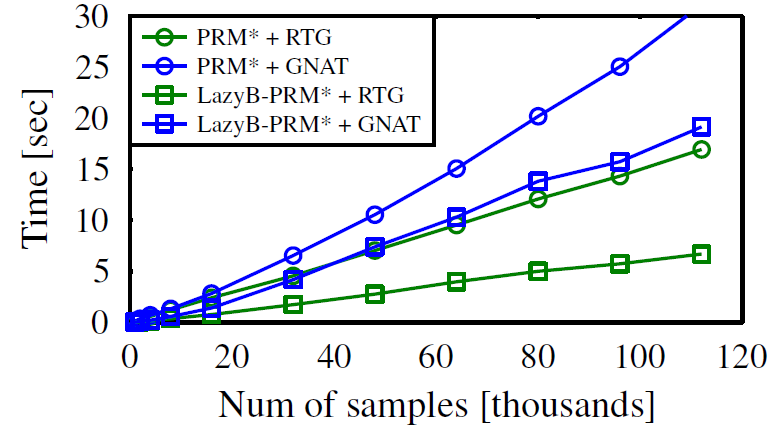}
   \vspace{-7pt}
   \caption{\sf \footnotesize Roadmap construction time in a three-dimensional Euclidean \Cs
   of a translating robot in the Z-tunnel scenario.}
   \label{fig:z-tunnel-const}
 \end{figure}

We first tested the construction time of the multi-query algorithms PRM*
and LazyB-PRM* on the Z-tunnel scenario\footnote{Scenario based on the Z tunnel scenario by the Parasol MP Group, CS Dept, Texas A\&M University 		 \url{https://parasol.tamu.edu/groups/amatogroup/benchmarks/mp/z_tunnel/}}
(Fig.~\ref{fig:corr}) in a three-dimensional Euclidean \Cs of a robot translating in space.
Fig.~\ref{fig:z-tunnel-const} reports on the construction time as a function of the number $n$ of samples in the roadmap.
One can clearly see that as $n$ grows, using RTG becomes more advantageous.
To test the quality of the roadmap obtained,
we performed a query for finding a path
from one side of the Z-tunnel to the other (see green and red robots in Fig.~\ref{fig:corr}). Both roadmaps yielded similar success rates in finding a solution.
\textVersion{%
We obtained similar results in higher dimensional \Css.}{}

 \begin{figure}[t]
  \centering
    \includegraphics[width=6.7cm]{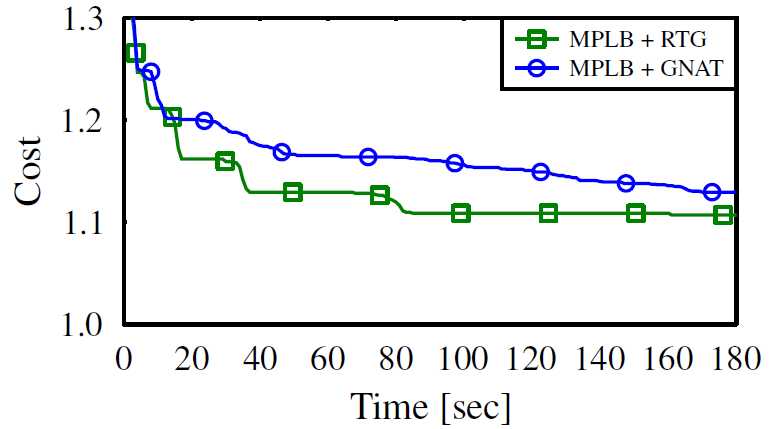}
   \vspace{-7pt}
   \caption{\sf \footnotesize Solution's cost vs. time in a six-dimensional MR-metric
   (non-Euclidean) \Cs
   of two translating robots in the 3D Grid scenario.
   The cost is normalized such that a cost of one denotes the optimal cost that may be obtained.}
 	\label{fig:3dgrid-MR-res}
 \end{figure}
\textVersion{}
{
 \begin{figure}[t]
  \centering
    \includegraphics[width=6.7cm]{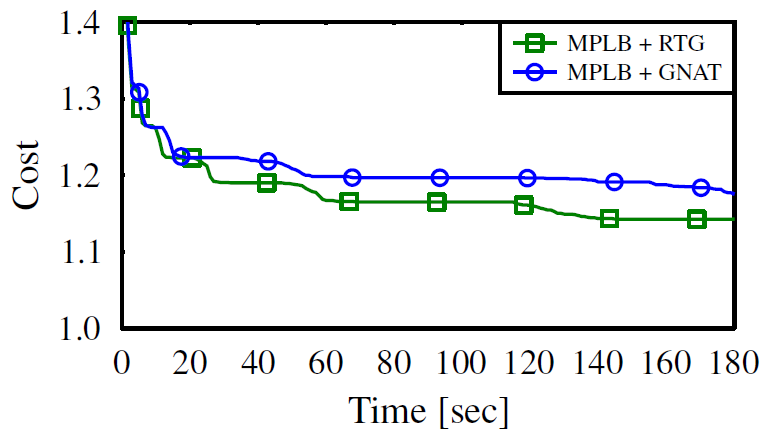}
   \vspace{-7pt}
   \caption{\sf \footnotesize Solution's cost vs. time in a six-dimensional Euclidean \Cs
   of two translating robots in the 3D Grid scenario.
   The cost is normalized such that a cost of one denotes the optimal cost that may be obtained.}
 	\label{fig:3dgrid-res}
 \end{figure}
}
 \begin{figure}[t]
  \centering
   	\includegraphics[width=6.7cm]{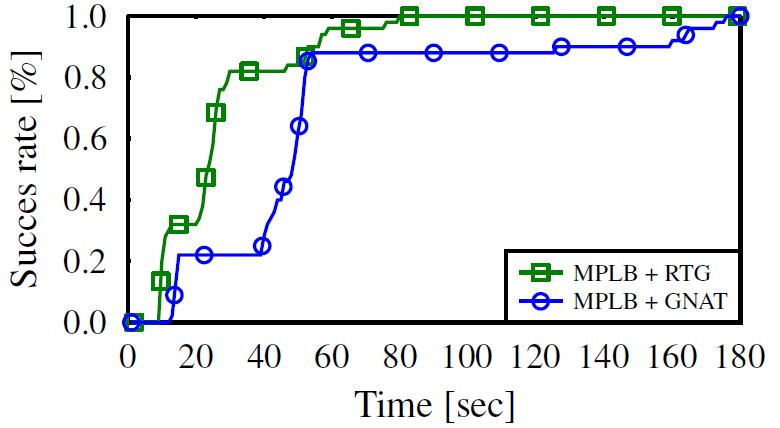}
      \vspace{-7pt}
   \caption{\sf \footnotesize Success rate to find a solution in a
   					six-dimensional Euclidean \Cs
   					of two translating robots in the Cubicles scenario.}
   \label{fig:cubicles-res}
\end{figure}
\textVersion{}
{\begin{figure}[t]
  \centering
   	\includegraphics[width=6.7cm]{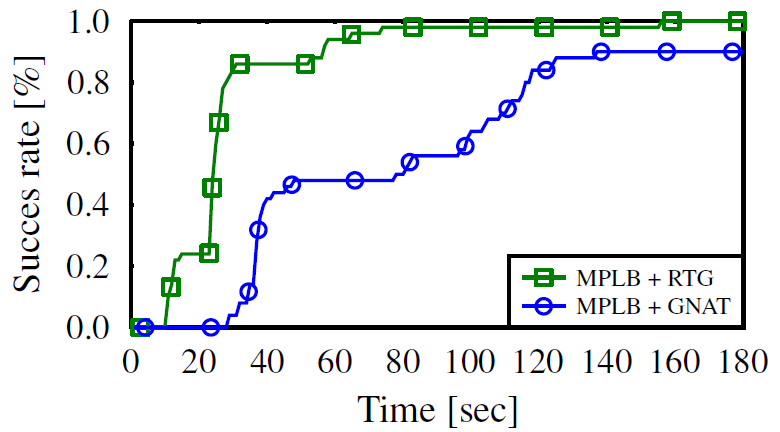}
      \vspace{-7pt}
   \caption{\sf \footnotesize Success rate to find a solution in a
   					six-dimensional MR-metric
                   (non-Euclidean) \Cs
   					of two translating robots in the Cubicles scenario.}
   \label{fig:cubicles-MR-res}
\end{figure}}

Next, we tested the single-query MPLB algorithm on the 3D Grid
scenario (Fig.~\ref{fig:3dgrid}) in a six-dimensional \Cs consisting of two translating robots in space.
The distance metric we used, which we refer to as MR-metric,
computes the sum of the distances that each robot travels.
Fig.~\ref{fig:3dgrid-MR-res} presents the quality of the solution obtained as a function of time.
One can see that RTG allows to find higher quality solutions faster than GNAT.
Even though the analysis of Aiger et al. regarding the
quality of RTG's results holds only in Euclidean
spaces, the MR-metric is more natural in multi-robot settings.
We repeated the same test for the Euclidean metric.
The results,
\textVersion{omitted due to lack of space}{presented in Fig.~\ref{fig:3dgrid-res}}, demonstrate similar trends to those presented for the MR-metric.

Finally, we report on the success rate of finding a solution in the Cubicles scenario\footnote{The Cubicles scenario is provided with the OMPL distribution.} (Fig.~\ref{fig:cubicles}).
We first performed the experiment for a six-dimensional Euclidean \Cs consisting of two translating robots.
The results, depicted in Fig.~\ref{fig:cubicles-res},
demonstrate that RTG allows to reduce the time to find an initial solution in complex~scenarios.
\textVersion{
Similar results were observed for the MR-metric. 
}{
Fig.~\ref{fig:cubicles-MR-res} presents similar results using the MR-metric.}

\section{Discussion and future work}
\label{sec:future}

Many possible enhancements can be applied to our \rtg implementation
in order to easily use it in sampling-based \MPd algorithms.
One obvious requirement is to automatically tune the two parameters
$c$ and $m$ defining the grid cell size and the number of constructed grids,
respectively.
%

Our RTG implementation is much faster when the flattened bit-array
is used for storing the pairs, however this structure has a quadratic space complexity.
Thus, a potential solution may introduce a hybrid data structure that for small number of samples uses the bit-array
and for large inputs uses the array of unordered sets.
Nevertheless, as the array of unordered sets is output sensitive in its memory consumption,
when the number of true neighboring pairs is very large,
cache faults occur and the program slows down drastically.
Thus, we seek to have an IO-efficient RTG implementation in order
to overcome these limitations.

Similar to the work in~\cite{AKR13}, one can implement RTG differently such that a
single $r$-nearest-neighbors query, as opposed to all-pairs $r$-\nns, is answered efficiently.
Such implementation should store all $m$ constructed grids.
At query phase, given a query point $q$, the $m$ cells containing $q$ are examined, and the set of neighboring points among all potential candidates within the cells is found.
For small values of $m$, this implementation may be efficient.

Another interesting extension is parallelizing the algorithm.
A possible approach may be
to partition the grid into
overlapping portions and run the algorithm on each portion separately.
Finally, we wish to devise an RTG variant applicable to non-Euclidean \Css, a much needed data structure in asymptotically-optimal sampling-based \MPd algorithms.

\section{Acknowledgments}
We wish to thank Dror Aiger for fruitful discussions regarding efficiently implementing the RTG algorithm.

\bibliographystyle{IEEEtran}
\bibliography{bibliography}

\end{document}